\documentclass[11pt,a4paper]{article}
\usepackage[resetfonts]{cmap} 
\usepackage[T2A,T1]{fontenc}
\usepackage[russian,english]{babel}
\usepackage[hyperref]{ranlp2021}
\usepackage{times}
\usepackage{latexsym}

\usepackage{microtype}
\usepackage{varioref}
\usepackage{amssymb}
\usepackage{graphicx}
\usepackage{url}
\usepackage{booktabs}
\usepackage{nicefrac}
\usepackage{subcaption}

\aclfinalcopy 

\def\aclpaperid{48} 

\newcommand\email[1]{{\small\texttt{#1}}}

\title{One Size Does Not Fit All: Finding the Optimal Subword Sizes\\ for FastText Models across Languages%
    \thanks{%
      First author's work was graciously funded by the South
      Moravian Centre for International Mobility as a part of the Brno
      Ph.D.\ talent project. Computational resources were supplied by
      the project ``e-Infrastruktura CZ'' (e-INFRA LM2018140) provided
      within the program Projects of Large Research, Development and
      Innovations Infrastructures.
    }
}

\iffalse 
  \author{Vít~Novotný,  Eniafe~Festus~Ayetiran,  Dalibor~Bačovský,\\  \textbf{Dávid~Lupták}, \textbf{Michal~Štefánik} and \textbf{Petr~Sojka}\\
  \email{\{witiko,ayetiran,456662,dluptak,stefanik.m,sojka\}@mail.muni.cz}
\else
\author{%
  Vít~Novotný\\
  Faculty of Informatics\\
  Masaryk University\\
  \email{witiko@mail.muni.cz} \\
\And
  Eniafe Festus Ayetiran\\
  Faculty of Informatics\\
  Masaryk University\\
  \email{ayetiran@mail.muni.cz} \\
\And
  Dalibor Bačovský\\
  Faculty of Informatics\\
  Masaryk University\\
  \email{456662@mail.muni.cz} \\
\AND
  Dávid Lupták\\
  Faculty of Informatics\\
  Masaryk University\\
  \email{dluptak@mail.muni.cz}~~~~ \\
\And
  Michal Štefánik\\
  Faculty of Informatics\\
  Masaryk University\\
  \email{stefanik.m@mail.muni.cz} \\
\And
  Petr Sojka\\
  Faculty of Informatics\\
  Masaryk University\\
  ~~\email{sojka@fi.muni.cz} \\
}
\fi

\begin{document}
\hyphenation{fast-text Graphi-Con}
\maketitle
\begin{abstract}
Unsupervised representation learning of words from large multilingual corpora is useful for downstream tasks such as word sense disambiguation, semantic text similarity, and information retrieval.
The representation precision of log-bilinear fastText models is mostly due to their use of subword information.

In previous work, the optimization of fastText's subword sizes has not been fully explored, and non-English fastText models were trained using subword sizes optimized for English and German word analogy tasks.

In our work, we find the optimal subword sizes on the English, German, Czech, Italian, Spanish, French, Hindi, Turkish, and Russian word analogy tasks.
We then propose a simple $n$-gram coverage model and we show that it predicts better-than-default subword sizes on the Spanish, French, Hindi, Turkish, and Russian word analogy tasks.

We show that the optimization of fastText's subword sizes matters and results in a 14\% improvement on the Czech word analogy task.
We also show that expensive parameter optimization can be replaced by a simple $n$-gram coverage model that consistently improves the accuracy of fastText models on the word analogy tasks by up to 3\% compared to the default subword sizes, and that it is within 1\% accuracy of the optimal subword sizes.

\end{abstract}

\section{Introduction}
\label{sec:introduction}

\citet{bojanowski2017enriching} have shown that taking word morphology into account is important for accurate continuous representations of words. However, they only show the optimal $n$-gram sizes on the German and English word analogy tasks~\cite[Section~5.5]{bojanowski2017enriching}. We continue their experiment by finding the optimal parameters on the Czech, Italian, Spanish, French, Hindi, Turkish, and Russian word analogy tasks and we show an up to 14\% improvement in accuracy compared to the default subword sizes.

Furthermore, we propose a cheap and simple $n$-gram coverage model that can suggest near-optimal subword sizes for under-resourced languages, where the optimal subword sizes are unknown. We train our $n$-gram coverage model on the English, German, Czech, and Italian word analogy tasks, and we show that it suggests subword sizes that improve the accuracy by up to $3\%$ on the Spanish, French, Hindi, Turkish, and Russian word analogy tasks and are within $1\%$ accuracy of the optimal subword sizes on average. To make it easy for others to reproduce and build upon our work, we have publicly released a reference implementation of our $n$-gram coverage model.\footnote{See \href{https://github.com/MIR-MU/fasttext-optimizer}
{https://github.com/MIR-MU/fasttext-optimizer}.}

The rest of the paper is structured as follows: In Section~\ref{sec:related-work}, we discuss the related work. In Section~\ref{sec:methods}, we discuss our methods and we propose our $n$-gram coverage model. In Section~\ref{sec:results}, we show and discuss our results. We conclude in Section~\ref{sec:conclusion} by summarizing our contribution. We outline the future work in Section~\ref{sec:future-work}.

\begin{table*}[tbh]
\centerline{\begin{minipage}{\textwidth}
\small\centering
\begin{subtable}{15.9em}
\centering\tabcolsep7dd
\vspace*{1.8em}
\begin{tabular}{ll}
\toprule
$n = 1$: & H, e, l, o, w, r, d \\
$n = 2$: & He, el, ll, lo, wo, or, rl, ld \\
$n = 3$: & Hel, ell, llo, wor, orl, rld \\
$n = 4$: & Hell, ello, worl, orld \\
$n = 5$: & Hello, world \\
\bottomrule
\end{tabular}
\caption{27 unique subwords in a corpus of two words: \emph{Hello} and \emph{world}}
\end{subtable}
\hfill
\begin{subtable}{12.4em}
\centering\tabcolsep9dd
\vspace*{1.8em}
\begin{tabular}{ll}
\toprule
$n = 1$: & $\nicefrac{7}{27} = 25.93\% $ \\
$n = 2$: & $\nicefrac{8}{27} = 29.63\% $ \\
$n = 3$: & $\nicefrac{6}{27} = 22.22\% $ \\
$n = 4$: & $\nicefrac{4}{27} = 14.81\% $ \\
$n = 5$: & $\nicefrac{2}{27} = \phantom{0}7.41\% $ \\
\bottomrule
\end{tabular}
\caption{Frequencies of unique subwords of size~$n$}
\end{subtable}
\hfill
\begin{subtable}{18em}
\centering \tabcolsep2.5dd
\begin{tabular}{cccccc}
\toprule
  & 1 & 2 & 3 & 4 & 5 \\
\midrule
1 & \phantom{0}25.93 & \phantom{0}55.55 & \phantom{0}77.77 & \phantom{0}92.59 &           100.00 \\
2 &                  & \phantom{0}29.63 & \phantom{0}51.85 & \phantom{0}66.66 & \phantom{0}74.07 \\
3 &                  &                  & \phantom{0}22.22 & \phantom{0}37.03 & \phantom{0}44.44 \\
4 &                  &                  &                  & \phantom{0}14.81 & \phantom{0}22.22 \\
5 &                  &                  &                  &                  & \phantom{00}7.41 \\
\bottomrule
\end{tabular}
\caption{$N$-gram coverages for the subword sizes $i$--$j$, where $1\leq i\leq j\leq 5$}
\end{subtable}
\end{minipage}}
\caption{%
	An example of $n$-gram coverages. We start in Subtable~(a) by producing all unique subwords of size less than $10$ from the corpus.
	In Subtable~(b), we compute the frequencies of unique subwords of different sizes.
	In Subtable~(c), we compute the $n$-gram coverages for various subword sizes $i$--$j$.
}
\label{table:ngram-coverages}
\end{table*}

\section{Related work}
\label{sec:related-work}

\citet{mikolov13efficient} described the \emph{Word2vec language model}, which uses a shallow neural network to learn continuous representations of words: \emph{word embeddings}.
They also produced the \emph{English word analogy task}, which tests how well word embeddings represent language regularities such as analogical relations (man is to woman what a king is to a queen), and evaluated Word2vec on their task.

\citet{berardi2015word, koper2015multilingual, svoboda2016new, cardellino2019spanish, gungor2017linguistic, korogodina2020evaluation} produced the \emph{Italian, German, Czech, Spanish, Turkish, and Russian word analogy tasks} for evaluating the performance of non-English word embeddings.
Their findings revealed that, despite the morphological complexity of the languages, Word2vec language models can generate semantically and syntactically meaningful word embeddings.
\looseness=-1

In order to take word morphology into account, \citet{bojanowski2017enriching} developed the \emph{fastText language model} based on Word2vec.
Their improvements consisted of representing each word as a sequence of \emph{subwords} with their own embeddings.
They evaluated their models on the English, German, Czech, and Italian word analogy tasks. They also showed the optimal subword sizes of fastText on the English and German word analogy tasks. However, they did not optimize the subword sizes of fastText on the Czech and Italian word analogy tasks and used subwords of size 3--6\footnote{For subword sizes, we adopt the notation of \citet{bojanowski2017enriching} and \cite{grave2018learning}. For example, subwords of size 3--6 are all subwords whose size is 3, 4, 5, or 6.}, which they described as ``an arbitrary choice''~\cite[Section~5.5]{bojanowski2017enriching}.

\citet{grave2018learning} produced the \emph{French and Hindi word analogy tasks}.
Furthermore, they also trained and publicly released fastText language models for 157~languages.
Like Bojanowski et al., they also neglected to optimize the subword sizes. Unlike Bojanowski et al., they used subwords of size~5--5 for all languages, noting that ``using character $n$-grams of size~5, instead of using the default range of 3--6, does not significantly decrease the accuracy (except for Czech).''~\cite[Section~4.3]{grave2018learning}




\section{Methods}
\label{sec:methods}

In this section, we describe our methods and propose our $n$-gram coverage model, which can be used to suggest subword sizes for a fastText model without expensive parameter optimization.

\subsection{Optimal subword sizes}
In the first part of our experiment, we train fastText language models on the English (22\,GiB), German (8.3\,GiB), Czech (1.2\,GiB), Italian (4.2\,GiB), Spanish (5.2\,GiB), French (7.4\,GiB), Hindi (0.57\,GiB), Turkish (0.72\,GiB), and Russian (9.9\,GiB) Wikipedia corpora.
We use subword sizes $i$--$j$ 
for all $i, j,$ where $1\leq i\leq j\leq 10$, and we report the accuracies and the optimal subword sizes $i$--$j$ on the English~\cite{mikolov13efficient}, German~\cite{koper2015multilingual}, Czech~\cite{svoboda2016new}, Italian~\cite{berardi2015word}, Spanish~\cite{cardellino2019spanish}, French and Hindi~\cite{grave2018learning}, Turkish~\cite{gungor2017linguistic}, and Russian\footnote{See \href{https://rusvectores.org/static/testsets/ru_analogy.txt}{https://rusvectores.org/static/testsets/ru\_analogy.txt}.} word analogy tasks.

\subsection{{\boldmath$N$}-gram coverage}
In the second part of our experiment, we compute and report the ratio between the frequencies of unique subwords of size $i$--$j$ and the frequencies of all unique subwords of size less than $10$ on the English, German, Czech, Italian, Spanish, French, Hindi, Turkish, and Russian Wikipedia corpora.
In the following text, we call this ratio the \emph{$n$-gram coverage}.
Table~\vref{table:ngram-coverages} shows how the $n$-gram coverage is computed by example.

\subsection{Suggested subword sizes}
In the third part of our experiment, we show that the $n$-gram coverage can be used to suggest subword sizes that are close to the optimal subword sizes on word analogy tasks.

For training, we compute \emph{the mean $n$-gram coverage for the optimal subword sizes} on the English, German, Czech, and Italian word analogy tasks.
For testing, we suggest subword sizes for the Spanish, French, Hindi, Turkish, and Russian word analogy tasks, so that the $n$-gram coverages for the suggested subword sizes on the testing word analogy tasks are \emph{the closest to the mean $n$-gram coverage for the optimal subword sizes} on the training word analogy tasks. Notice that the suggested subword sizes are not based on the optimal subword sizes for the testing word analogy tasks.

After the performance estimation, we fold the training and testing word analogy tasks and we compute the mean $n$-gram coverage for the optimal subword sizes on all word analogy tasks (English, German, Czech, Italian, Spanish, French, Hindi, Turkish, and Russian). This means $n$-gram coverage can be used in applications of fastText to suggest the optimal subword sizes without expensive parameter optimization.

\subsection{Language distances}

In the final part of our experiment, we interpret suggested subword sizes as \emph{two-dimensional vectors} and use \emph{the Euclidean distance} to measure distances between languages. To see if our language distance measure represents interpretable linguistic phenomena, we compare it to the typological, geographical, and phylogenetic language distance measures of \citet{littell2017uriel}:
\begin{description}
\item[Typological] \citeauthor{littell2017uriel} define three typological language distance measures: \emph{syntactic}, \emph{phonological}, and \emph{inventory}. Each distance measure is defined as the cosine distances between different feature vectors:
\end{description}
\begin{itemize}
\item The \emph{syntactic} features describe the sentence structure of a language and have been adapted from the World Atlas of
Language Structures (WALS), Syntactic Structures
of World Languages, and Ethnologue databases.
\item The \emph{phonological} features describe the structure of the sound and sign systems of a language and have been adapted from the WALS and Ethnologue databases.
\item The \emph{inventory} features describe the presence or absence of distinctive speech sounds in the sound system of a language and have been adapted from the PHOIBLE database.
\end{itemize}
\begin{description}
\item[Geographical] The \emph{geographical} language distance measure is defined as the cosine distance between feature vectors, where the features have been adapted from declarations of language location in the Glottolog, WALS, and SSWL databases.
\item[Phylogenetic] The \emph{phylogenetic} language distance measure is defined as the cosine distance between feature vectors, where the features correspond to the shared membership in language families, according to the world language family tree in the Glottolog database.
\end{description}

To compare our language distance measure with the language distance measures of \citeauthor{littell2017uriel}, we compute and report the Pearson's correlation coefficient ($r$) between the distance measures.

\subsection{Implementation details}
We reproduce the experimental setup of \citet[Section~4]{bojanowski2017enriching}: skip-gram architecture, hash table bucket size $2 \cdot 10^6$, 300 vector dimensions, negative sampling loss with 5~negative samples, initial learning rate 0.05 with a linear decay to zero, sampling threshold $10^{-4}$, window size~5, and 5~epochs.

Like \citet{bojanowski2017enriching}, we use a reduced vocabulary of the $2\cdot 10^5$ most frequent words to solve word analogies.
We use the implementation of word analogies in Gensim~\cite{rehurek2010software}, which uses Unicode upper-casing in the \texttt{en\_US.UTF-8} locale for caseless matching.

To compute Pearson's $r$ between two language distance measures, we use the Representational Similarity Analysis (RSA) framework of \citet{kriegeskorte2008representational, chrupala2019correlating}: we produce two matrices of all pairwise distances between 282 Wikipedia languages\footnote{For six out of the 288~Wikipedia languages, \citeauthor{littell2017uriel} did not provide feature vectors: Bhojpuri (\texttt{bh}), Emilian-Romagnol (\texttt{eml}), Western Armenian (\texttt{hyw}), Nahuatl (\texttt{nah}), Simple English (\texttt{simple}), and Sakizaya (\texttt{szy}).} and we compute Pearson's~$r$ between the upper-triangulars, excluding the diagonals.

To make it easy for others to reproduce and build upon our work, we have published a reference implementation of our $n$-gram coverage model, which suggests subword sizes for fastText models.\footnote{See \href{https://github.com/MIR-MU/fasttext-optimizer}{https://github.com/MIR-MU/fasttext-optimizer}.} The reference implementation contains pre-computed subword frequencies for 288 Wikipedia languages, which makes the suggestions instantaneous.

\begin{table*}[tbhp]
\hspace*{-0.05\textwidth}
\begin{minipage}{1.1\textwidth}
\begin{subtable}{0.25\textwidth}
\tabcolsep3.5dd
\centering
\begin{tabular}{cccccc}
\toprule
& 2 & 3 & 4 & \textbf{5} & 6 \\
\midrule
2                        & 73 & 73 & 73 & 74              & 74 \\
3 &               & 74 & 75 & 76 & 75 \\
\textbf{4}              &               &               & 76 & \textbf{76}    & 76 \\
5                        &               &               &               & 76              & 76 \\
6                        &               &               &               &                            & 75 \\
\bottomrule
\end{tabular}
\caption{English}
\label{table:accuracies-en}
\end{subtable}%
\begin{subtable}{0.25\textwidth}
\tabcolsep3.5dd
\centering
\begin{tabular}{cccccc}
\toprule
& 2 & 3 & 4 & 5 & \textbf{6} \\
\midrule
2           & 51 & 52 & 54 & 56 & 57 \\
3 &               & 55 & 56 & 57 & 58 \\
4           &               &               & 57 & 58 & 59 \\
5           &               &               &               & 59 & 60 \\
\textbf{6} &               &               &               &               & \textbf{61} \\
\bottomrule
\end{tabular}
\caption{German}
\label{table:accuracies-de}
\end{subtable}%
\begin{subtable}{0.25\textwidth}
\tabcolsep3.5dd
\centering
\begin{tabular}{ccccccc}
\toprule
& 2 & 3 & \textbf{4} & 5 & 6 \\
\midrule
\textbf{1}    & 44 & 58 & \textbf{60}    & 60 & 58 \\
2 & 41 & 57 & 59 & 58 & 57 \\
3    &               & 53 & 56              & 54 & 55 \\
4              &               &               & 49              & 52 & 49 \\
5              &               &               &                            & 46 & 46 \\
\bottomrule
\end{tabular}
\caption{Czech}
\label{table:accuracies-cs}
\end{subtable}%
\begin{subtable}{0.25\textwidth}
\tabcolsep3.5dd
\centering
\begin{tabular}{ccccccc}
\toprule
& 2 & 3 & 4 & \textbf{5} & 6 \\
\midrule
1 & 44 & 50 & 53 & 54 & 53 \\
\textbf{2}    & 46 & 51 & 53 & \textbf{54}    & 52 \\
3    &               & 51 & 53 & 53 & 53 \\
4 &               &               & 53 & 53 & 52 \\
5              &               &               &               & 53              & 52 \\
\bottomrule
\end{tabular}
\caption{Italian}
\label{table:accuracies-it}
\end{subtable}\par\medskip
\begin{subtable}{0.2\textwidth}
\tabcolsep2.2dd
\centering
\begin{tabular}{ccccccc}
\toprule
& 2 & 3 & 4 & \textbf{5} & 6 \\
\midrule
2 & 51 & 53 & 55 & 55 & 55 \\
3 &    & 55 & 57 & 57 & 56 \\
4 &    &    & 57 & 57 & 57 \\
\textbf{5} &    &    &    & \textbf{57} & 57 \\
6 &    &    &    &    & 57 \\
\bottomrule
\end{tabular}
\caption{Spanish}
\label{table:accuracies-es}
\end{subtable}%
\begin{subtable}{0.2\textwidth}
\tabcolsep2.2dd
\centering
\begin{tabular}{ccccccc}
\toprule
& 2 & 3 & 4 & 5 & \textbf{6} \\
\midrule
2 &  63 &  63 &  65 &  67 &  67 \\
3 &     &  66 &  66 &  67 &  68 \\
4 &     &     &  68 &  68 &  69 \\
5 &     &     &     &  69 &  69 \\
\textbf{6} &     &     &     &     &  \textbf{70} \\
\bottomrule
\end{tabular}
\caption{French}
\label{table:accuracies-fr}
\end{subtable}%
\begin{subtable}{0.2\textwidth}
\tabcolsep2.2dd
\centering
\begin{tabular}{ccccccc}
\toprule
& \textbf{2} & 3 & 4 & 5 & 6 \\
\midrule
1 &  15 &  15 &  14 &  14 &  14 \\
\textbf{2} &  \textbf{17} &  15 &  14 &  13 &  14 \\
3 &     &  16 &  13 &  13 &  12 \\
4 &     &     &  13 &  12 &  12 \\
5 &     &     &     &  12 &  11 \\
\bottomrule
\end{tabular}
\caption{Hindi}
\label{table:accuracies-hi}
\end{subtable}%
\begin{subtable}{0.2\textwidth}
\tabcolsep2.2dd
\centering
\begin{tabular}{ccccccc}
\toprule
& 2 & \textbf{3} & 4 & 5 & 6 \\
\midrule
1 &  32 &  37 &  38 &  38 &  37 \\
2 &  34 &  39 &  39 &  39 &  38 \\
\textbf{3} &     &  \textbf{40} &  39 &  39 &  38 \\
4 &     &     &  37 &  38 &  37 \\
5 &     &     &     &  36 &  35 \\
\bottomrule
\end{tabular}
\caption{Turkish}
\label{table:accuracies-tr}
\end{subtable}%
\begin{subtable}{0.2\textwidth}
\tabcolsep2.2dd
\centering
\begin{tabular}{ccccccc}
\toprule
& 2 & 3 & 4 & 5 & \textbf{6} \\
\midrule
2 &  46 &  43 &  46 &  50 &  51 \\
3 &     &  46 &  47 &  50 &  52 \\
4 &     &     &  51 &  51 &  52 \\
\textbf{5} &     &     &     &  52 &  \textbf{53} \\
6 &     &     &     &     &  53 \\
\bottomrule
\end{tabular}
\caption{Russian}
\label{table:accuracies-ru}
\end{subtable}%
\end{minipage}%
\caption{Accuracies on English, German, Czech, Italian, Spanish, French, Hindi, Turkish, and Russian word analogy tasks. Optimal subword sizes for the different word analogy tasks are \textbf{bold}: 4--5 for English, 6--6 for German, 1--4 for Czech, 2--5 for Italian, 5--5 for Spanish, 6--6 for French, 2--2 for Hindi, 3--3 for Turkish, and 5--6 for Russian. Our training and testing word analogy tasks are shown on separate lines.}
\label{table:accuracies}
\end{table*}

\begin{table*}[tbhp]
\hspace*{-0.05\textwidth}
\begin{minipage}{1.1\textwidth}
\begin{subtable}{0.33\textwidth}
\tabcolsep2dd
\centering
\begin{tabular}{ccccccc}
\toprule
& 2 & 3 & 4 & \textbf{5} & 6 \\
\midrule
2 &  \phantom{0}0.26 &  \phantom{0}0.75 &  \phantom{0}1.72 &  \phantom{0}4.51 &  10.50 \\
3  &   &  \phantom{0}0.49 & \phantom{0}1.45 & \phantom{0}4.25 & 10.24 \\
\textbf{4}  &   &   &  \phantom{0}0.97 & \textbf{\phantom{0}3.76} &  \phantom{0}9.75 \\
5 &   &   &   &  \phantom{0}2.79 &  \phantom{0}8.78 \\
6 &   &   &   &   &  \phantom{0}5.99 \\
\bottomrule
\end{tabular}
\caption{English}
\label{table:coverages-en}
\end{subtable}%
\begin{subtable}{0.33\textwidth}
\tabcolsep2dd
\centering
\begin{tabular}{ccccccc}
\toprule
& 2 & 3 & 4 & 5 & \textbf{6} \\
\midrule
2 &  \phantom{0}0.08 &  \phantom{0}0.25 &  \phantom{0}0.85 &  \phantom{0}2.68 &  \phantom{0}6.87 \\
3 &   &  \phantom{0}0.17 &  \phantom{0}0.77 &  \phantom{0}2.60 &  \phantom{0}6.79 \\
4 &   &   &  \phantom{0}0.60 &  \phantom{0}2.43 &  \phantom{0}6.62 \\
5 &   &   &   &  \phantom{0}1.83 &  \phantom{0}6.02 \\
\textbf{6} &   &   &   &   &  \textbf{\phantom{0}4.19} \\
\bottomrule
\end{tabular}
\caption{German}
\label{table:coverages-de}
\end{subtable}%
\begin{subtable}{0.33\textwidth}
\tabcolsep2dd
\centering
\begin{tabular}{ccccccc}
\toprule
& 2 & 3 & \textbf{4} & 5 & 6 \\
\midrule
\textbf{1} & \phantom{0}0.21 &  \phantom{0}0.82 &  \textbf{\phantom{0}3.28} &  10.40 &  23.23 \\
2 &  \phantom{0}0.18 &  \phantom{0}0.80 &  \phantom{0}3.25 &  10.37 &  23.20 \\
3 &   &  \phantom{0}0.61 &  \phantom{0}3.07 &  10.19 &  23.02 \\
4 &   &   &  \phantom{0}2.46 &  \phantom{0}9.58 &  22.40 \\
5 &   &   &   &  \phantom{0}7.12 &  19.95 \\
\bottomrule
\end{tabular}
\caption{Czech}
\label{table:coverages-cs}
\end{subtable}\par\medskip
\begin{subtable}{0.33\textwidth}
\tabcolsep2dd
\centering
\begin{tabular}{ccccccc}
\toprule
& 2 & 3 & 4 & \textbf{5} & 6 \\
\midrule
\textbf{2} &  \phantom{0}0.14 &  \phantom{0}0.44 &  \phantom{0}1.34 &  \textbf{\phantom{0}3.81} &  \phantom{0}8.92 \\
3 &   &  \phantom{0}0.30 &  \phantom{0}1.20 &  \phantom{0}3.67 &  \phantom{0}8.78 \\
4 &   &   &  \phantom{0}0.90 &  \phantom{0}3.37 &  \phantom{0}8.48 \\
5 &   &   &   &  \phantom{0}2.47 &  \phantom{0}7.58 \\
6 &   &   &   &   &  \phantom{0}5.11 \\
\bottomrule
\end{tabular}
\caption{Italian}
\label{table:coverages-it}
\end{subtable}%
\begin{subtable}{0.33\textwidth}
\tabcolsep2dd
\centering
\begin{tabular}{ccccccc}
\toprule
& 2 & 3 & 4 & \textbf{5} & 6 \\
\midrule
2 &   \phantom{0}0.25 &   \phantom{0}0.78 &   \phantom{0}2.35 &   \phantom{0}6.67 & 15.44 \\
3 &        &   \phantom{0}0.53 &   \phantom{0}2.10  &   \phantom{0}6.42 & 15.19 \\
4 &        &        &   \phantom{0}1.57 &   \phantom{0}5.89 & 14.66 \\
\textbf{5} &        &        &        &   \textbf{\phantom{0}4.32} & 13.09 \\
6 &        &        &        &        &  \phantom{0}8.77 \\
\bottomrule
\end{tabular}
\caption{Spanish}
\label{table:coverages-es}
\end{subtable}%
\begin{subtable}{0.33\textwidth}
\tabcolsep2dd
\centering
\begin{tabular}{ccccccc}
\toprule
& 2 & 3 & 4 & 5 & \textbf{6} \\
\midrule
2 &   \phantom{0}0.28 &   \phantom{0}0.85 &   \phantom{0}2.51 &   \phantom{0}7.00    & 16.21 \\
3 &        &   \phantom{0}0.57 &   \phantom{0}2.23 &   \phantom{0}6.73 & 15.93 \\
4 &        &        &   \phantom{0}1.66 &   \phantom{0}6.16 & 15.36 \\
5 &        &        &        &   \phantom{0}4.50  & 13.70  \\
\textbf{6} &        &        &        &        &  \textbf{\phantom{0}9.20}  \\
\bottomrule
\end{tabular}
\caption{French}
\label{table:coverages-fr}
\end{subtable}\par\medskip
\begin{subtable}{0.33\textwidth}
\tabcolsep2dd
\centering
\begin{tabular}{ccccccc}
\toprule
& \textbf{2} & 3 & 4 & 5 & 6 \\
\midrule
1 &   \phantom{0}0.70  &   \phantom{0}2.79 &   \phantom{0}8.15 &  18.01 & 30.59 \\
\textbf{2} &   \textbf{\phantom{0}0.57} &   \phantom{0}2.66 &   \phantom{0}8.03 &  17.89 & 30.46 \\
3 &        &   \phantom{0}2.09 &   \phantom{0}7.46 &  17.32 & 29.89 \\
4 &        &        &   \phantom{0}5.36 &  15.22 & 27.80 \\
5 &        &        &        &   \phantom{0}9.86 & 22.44 \\
\bottomrule
\end{tabular}
\caption{Hindi}
\label{table:coverages-hi}
\end{subtable}%
\begin{subtable}{0.33\textwidth}
\tabcolsep2dd
\centering
\begin{tabular}{ccccccc}
\toprule
& 2 & \textbf{3} & 4 & 5 & 6 \\
\midrule
1 &   \phantom{0}0.31 &   \phantom{0}1.08 &   \phantom{0}3.72 &  10.66 & 22.49 \\
2 &   \phantom{0}0.27 &   \phantom{0}1.04 &   \phantom{0}3.68 &  10.63 & 22.45 \\
\textbf{3} &        &   \textbf{\phantom{0}0.77} &   \phantom{0}3.41 &  10.35 & 22.18 \\
4 &        &        &   \phantom{0}2.64 &   \phantom{0}9.59 & 21.41 \\
5 &        &        &        &   \phantom{0}6.95 & 18.77 \\
\bottomrule
\end{tabular}
\caption{Turkish}
\label{table:coverages-tr}
\end{subtable}%
\begin{subtable}{0.33\textwidth}
\tabcolsep2dd
\centering
\begin{tabular}{ccccccc}
\toprule
& 2 & 3 & 4 & 5 & \textbf{6} \\
\midrule
2 &   \phantom{0}0.17 &   \phantom{0}0.61 &   \phantom{0}2.25 &   \phantom{0}7.13 & 16.57 \\
3 &        &   \phantom{0}0.43 &   \phantom{0}2.08 &   \phantom{0}6.96 & 16.40  \\
4 &        &        &   \phantom{0}1.64 &   \phantom{0}6.52 & 15.96 \\
\textbf{5} &        &        &        &   \phantom{0}4.88 & \textbf{14.32} \\
6 &        &        &        &        &  \phantom{0}9.44 \\
\bottomrule
\end{tabular}
\caption{Russian}
\label{table:coverages-ru}
\end{subtable}%
\end{minipage}%
\caption{The $n$-gram coverages for English, German, Czech, Italian, Spanish, French, Hindi, Turkish, and Russian. The $n$-gram coverages for the optimal subword sizes on the different word analogy tasks are \textbf{bold}: $3.76\%$ for English, $4.19\%$ for German, $3.28\%$ for Czech, $3.81\%$ for Italian, $4.32\%$ for Spanish, $9.20\%$ for French, $0.57\%$ for Hindi, $0.77\%$ for Turkish, and $14.32\%$ for Russian.}
\label{table:coverages}
\end{table*}

\begin{table*}[tb]
\tabcolsep12dd
\centering
\begin{tabular}{lllll}
\toprule
Language & \multicolumn{2}{l}{Default subword sizes} & Suggested subword sizes & Optimal subword sizes \\
& \multicolumn{1}{p{1.3cm}}{3--6} & \multicolumn{1}{p{1.3cm}}{5--5} & \\
\midrule
Spanish & \textit{57.00} & \textbf{57.60} & \textbf{57.60 (5--5)} & \textbf{57.60 (5--5)} \\
French  & 68.38 & \textit{69.33} & \textit{69.33 (5--5)} & \textbf{69.60 (6--6)} \\
Hindi   & 12.87 & 12.10 & \textit{15.03 (1--3)} & \textbf{16.95 (2--2)} \\
Turkish & 38.04 & 36.10 & \textit{38.34 (1--4)} & \textbf{39.51 (3--3)} \\
Russian & 51.89 & \textit{52.51} & \textit{52.51 (5--5)} & \textbf{52.75 (5--6)} \\
\bottomrule
\end{tabular}
\caption{Accuracies on the Spanish, French, Hindi, Turkish, and Russian word analogy tasks using the default subword sizes of \citeauthor{bojanowski2017enriching} (3--6) and \citeauthor{grave2018learning} (5--5), the subword sizes suggested by $n$-gram coverage, and the optimal subword sizes. Best accuracies for each language are \textbf{bold}, second best are in \textit{italics}.}
\label{table:accuracies-test}
\end{table*}

\section{Results}
\label{sec:results}

In this section, we show and discuss the optimal subword sizes, accuracies, and $n$-gram coverages on the English, German, Czech, Italian, Spanish, French, Hindi, Turkish, and Russian word analogy tasks. We also show that the $n$-gram coverage can be used to suggest subword sizes that are close to the optimal subword sizes.

\subsection{Optimal subword sizes}
In Table~\vref{table:accuracies}, we show the accuracies and the optimal subword sizes on the English, German, Czech, Italian, Spanish, French, Hindi, Turkish, and Russian word analogy tasks. The optimal subword sizes 4--5 for English and 6--6 for German reproduce and confirm the results of~\citet[Section~5.2]{bojanowski2017enriching}.

The optimal subword sizes for English (4--5), Italian (2--5), Spanish (5--5), French (6--6), and Russian (5--6) word analogy tasks are equal or within 1\% accuracy of the default subword sizes suggested by \citeauthor{bojanowski2017enriching} (3--6) and \citeauthor{grave2018learning} (5--5). In contrast, we see an improvement of up to 14\% for Czech (1--4), 5\% for Hindi (2--2), 4\% for Turkish (3--3), and 3\% for German (6--6).

To understand these differences, we look to the linguistic typology of languages: Czech, Hindi, and Turkish are synthetic languages and benefit from short subwords that represent morphemes. German and Russian are also synthetic, but the long compound nouns in German and the use of separate characters for yers (\foreignlanguage{russian}{ъ} and \foreignlanguage{russian}{ь}) in Russian make both languages benefit from longer subwords.

\subsection{{\boldmath$N$}-gram coverage}

In Table~\vref{table:coverages}, we show the $n$-gram coverages for English, German, Czech, Italian, Spanish, French, Hindi, Turkish, and Russian.

The mean $n$-gram coverage for the optimal subword sizes on the training word analogy tasks (English, German, Czech, and Italian), which we use to suggest subword sizes for the testing word analogy tasks (Spanish, French, Hindi, Turkish, and Russian), is $3.76\%$.
The mean $n$-gram coverage for the optimal subword sizes on all word analogy tasks, which can be used in applications of fastText to suggest the optimal subword sizes, is $4.91\%$.

\subsection{Suggested subword sizes}

In Table~\vref{table:accuracies-test}, we compare the accuracies on the testing word analogy tasks (Spanish, French, Hindi, Turkish, and Russian) using the default subword sizes of \citeauthor{bojanowski2017enriching} (3--6) and \citeauthor{grave2018learning} (5--5), the subword sizes suggested by the $n$-gram coverage, and the optimal subword sizes.

Using the suggested subword sizes is never worse than using the default subword sizes. For Hindi and Turkish, the suggested subword sizes always improve the accuracy: by $2.58\%$ on average compared to the weaker default subword sizes and by $1.23\%$ on average compared to the stronger default subword sizes. For Spanish, French, and Russian, the suggested subword sizes equal the default subword sizes of Grave et al. and they improve the accuracy by $0.72\%$ on average compared to the default subword sizes of Bojanowski et al.

For Spanish, the optimal subword sizes equal the suggested subword sizes. For French, Hindi, Turkish, and Russian, the optimal subword sizes improve the accuracy by only $0.90\%$ on average compared to the suggested subword sizes, whereas they improve the accuracy by $2.59\%$ on average compared to the weaker default subword sizes and by $1.52\%$ on average compared to the stronger default subword sizes.

\begin{figure}[tb!]
    \includegraphics[width=\columnwidth]{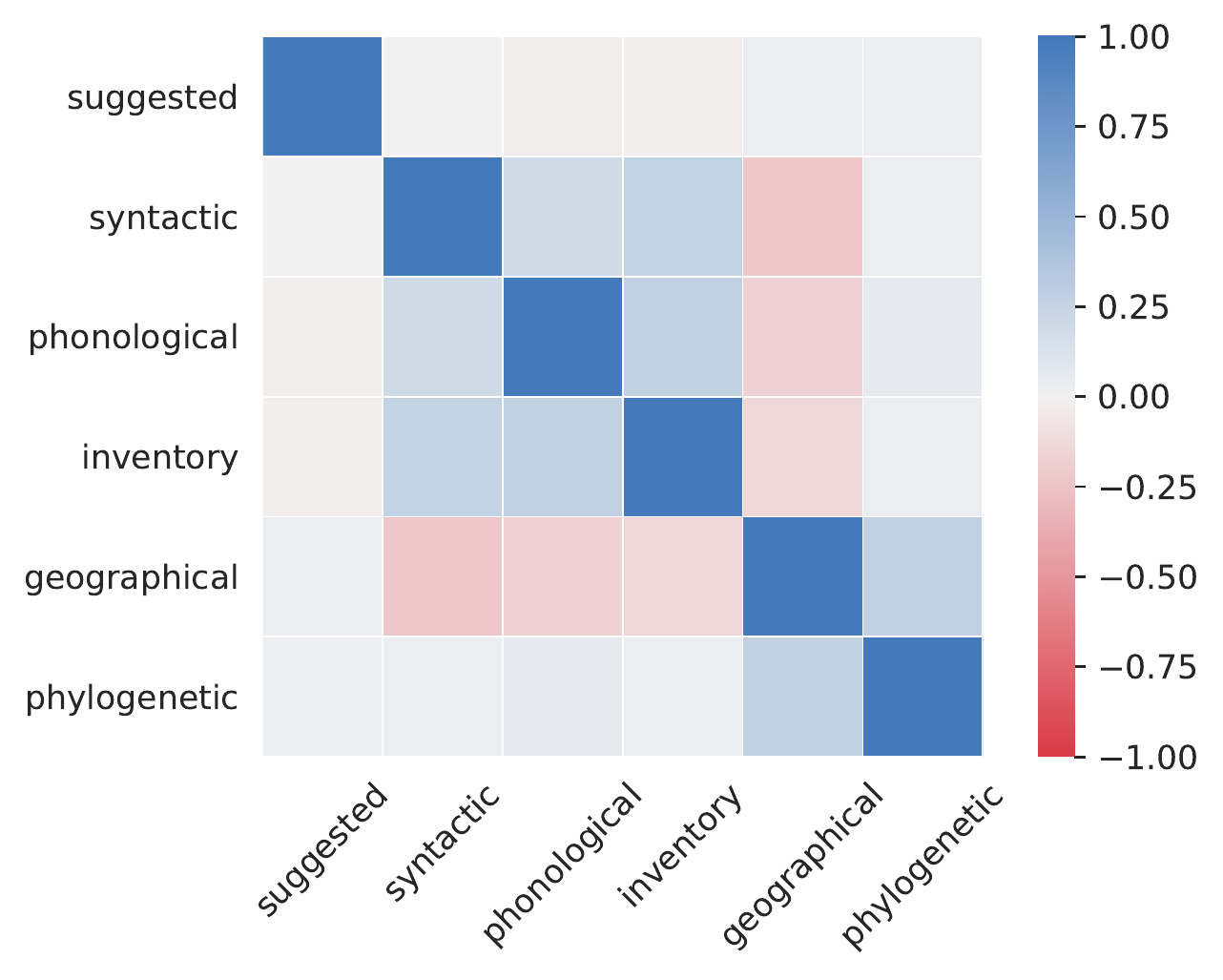}
    \caption{Pearson's correlation coefficients ($r$) between our language distance measure (\emph{suggested}) as well as the typological (\emph{syntactic}, \emph{phonological}, and \emph{inventory}), geographical, and phylogenetic language distance measures of \citet{littell2017uriel}. Best viewed in color.}
    \label{fig:language-distances}
\end{figure}

\subsection{Language distances}

In Figure~\vref{fig:language-distances}, we show Pearson's~$r$ between pairs of different language distance measures: 
our language distance measure, which is based on the Euclidean distance between our suggested subword sizes, as well as the typological, geographical, and phylogenetic language distance measures of \citeauthor{littell2017uriel}

Pearson's $r$ between our language distance measure and the language distance measures of \citeauthor{littell2017uriel} range between $-0.03$ (\emph{phonological}) and $0.03$ (\emph{geographical}). Since the absolute values of Pearson's $r$ are consistently smaller than random, our language distance measure does not either correlate or anti-correlate with the other language distance measures. This is because our suggested subword sizes are based on latent data-driven features of text, which complement the hand-crafted linguistic features of \citeauthor{littell2017uriel}

\section{Conclusion}
\label{sec:conclusion}
Subword sizes have a profound impact on the accuracy of fastText language models and their word embeddings. However, they are expensive to optimize on large corpora.

In this work, we showed the optimal subword sizes for Czech, Italian, Spanish, French, Hindi, Turkish, and Russian fastText language models, we confirmed prior optimal subword sizes reported for English and German, and we showed that the optimization of subword sizes improves the accuracy of fastText on word analogy tasks by up to 14\% compared to the default subword sizes. Our optimal subword sizes can be used in applications of fastText as the new default.

Furthermore, we proposed a cheap and simple $n$-gram coverage model that consistently improves the accuracy of fastText models on the word analogy tasks by up to 3\% compared to the default subword sizes, and that it is within 1\% accuracy of the optimal subword sizes on average. Subword sizes suggested by our $n$-gram coverage model can be used in applications of fastText as the new default for under-resourced languages, where the optimal subword sizes are unknown.

\section{Future work}
\label{sec:future-work}

Although the word analogy intrinsic task is a convenient proxy for the usefulness of fastText word embeddings, \citet{ghannay2016word,chiu2016intrinsic,rogers2018whats} show that it is no substitute for actual extrinsic end tasks.
In future work, we will evaluate our $n$-gram coverage model on extrinsic tasks.
\looseness=-1

In recent machine translation models~\cite{vaswani2017attention}, text is tokenized into words and subwords using word-piece~\cite{yonghui2016google} and byte-pair~\cite{sennrich2016neural} models. Our experiments suggest that we can remove the subword size parameter from fastText models and draw subwords from byte-pair models with little adverse effect on the word analogy accuracy. In future work, we will evaluate the use of word-piece and byte-pair models for subword selection in fastText models both on the intrinsic word analogy task and on other extrinsic tasks.

\bibliographystyle{acl_natbib}
\bibliography{main}


\end{document}